%% file: root.tex
\documentclass[a4paper,conference]{IEEEtran}
\input{packages.tex}

\begin{document}


\title{Multi-task multiple kernel
machines for personalized pain recognition from functional near-infrared spectroscopy brain signals}

\author{\IEEEauthorblockN{Daniel Lopez-Martinez$^{1,2}$,
Ke Peng$^{3,4}$,
Sarah C. Steele$^{3,4,5}$,
Arielle J. Lee$^{3,4,5}$,
David Borsook$^{3,4,5}$,
Rosalind Picard$^{2}$} \\
\IEEEauthorblockA{$^1$Harvard-MIT Health Sciences and Technology, Massachusetts Institute of Technology \tt{dlmocdm@mit.edu}}
\IEEEauthorblockA{$^2$Affective Computing group, MIT Media Lab, Massachusetts Institute of Technology}
\IEEEauthorblockA{$^3$Center for Pain and the Brain, Harvard Medical School} 
\IEEEauthorblockA{$^4$Department of Anesthesiology, Boston Children's Hospital}
\IEEEauthorblockA{$^5$Anthinoula A. Martinos Center for Biomedical Imaging, Department of Radiology, Massachusetts General Hospital}
}
\maketitle

\begin{abstract}
Currently there is no validated objective measure of pain. Recent neuroimaging studies have explored the feasibility of using functional near-infrared spectroscopy (fNIRS) to measure alterations in brain function in evoked and ongoing pain. In this study, we applied multi-task machine learning methods to derive a practical algorithm for pain detection derived from fNIRS signals in healthy volunteers exposed to a painful stimulus. Especially, we employed multi-task multiple kernel learning to account for the inter-subject variability in pain response. Our results support the use of fNIRS and machine learning techniques in developing objective pain detection, and also highlight the importance of adopting personalized analysis in the process.
\end{abstract}

\IEEEpeerreviewmaketitle

\section{Introduction}

Pain is a subjective, unpleasant sensory and emotional experience associated with actual or potential tissue damage \cite{DeCWilliams2016}. While current assessment of pain largely relies on the self-report of an individual \cite{Younger2009}, the development of an objective, automatic detection/measure of pain may be useful in many clinical applications including patients who are unconscious during tissue damage (viz., surgical patients) or who have difficulties in verbal or motor expressions (such as neonates or patients having cognitive impairments).  Such approaches, if successful may not only detect pain, but may provide for a more rational therapeutic intervention.

Previous work utilizing objective markers for pain recognition mainly employs approaches that focus on a subject’s physiological signals (such as heart rate and skin conductance) \cite{dlm_MTL_2017,dlmNIPS2017} or facial expression \cite{LopezMartinez2017c,dlmNIPS2017}. However, recently several studies have shown the potential of using functional near-infrared spectroscopy (fNIRS) in deriving a brain marker of pain perception \cite{Aasted2016,Gelinas2010,Yucel2015}. FNIRS is a noninvasive neuroimaging technique that is able to provide continuous recordings of cortical hemodynamic signals in terms of the concentration changes of oxygenated hemoglobin (HbO) and deoxygenated hemoglobin (HbR) \cite{Obrig2014}. 
A recent study has shown the feasibility of combining fNIRS and machine-learning techniques to perform post-hoc classification and clustering of high pain and low pain stimuli \cite{Pourshoghi2016}. However, a central difficulty in such approaches is that inter-subject variability in pain responses limits the ability for automated methods to generalize across people \cite{Dube2009,dlmNIPS2017}. Furthermore, the interpretability of the learned classifier with respect to brain functions and anatomy is an important, non-trivial, but usually difficult issue.
 
In this work, we employ multi-task learning (MTL) \cite{Caruana1997} to account for  individual differences in pain responses. This is a learning paradigm in machine learning in which models for several related tasks, each of which has limited training samples but share a latent structural relation, are trained simultaneously while using a shared representation to help improve the generalization performance of all the tasks. Here, we define each task as  pain detection  on a group of subjects that share similar pain responses. Task assignment is done by means of spectral clustering \cite{Planck2006}, although other clustering methods may be used.  MTL is then able to use what is learned for each task to help other tasks be learned better. Specifically, we employ a multi-task multiple kernel (MT-MKL) machine \cite{Kandemir2014} as our MTL algorithm. MT-MKL also offers the advantage that it allows training models on multiple fNIRS channels while automatically revealing which channels are useful for solving the tasks. Automatically learning the fNIRS channel importance is especially useful for interpreting the resulting classifier in terms of brain anatomy and functions.

The main contributions of this work are: (1) we present an interpretable machine learning approach to identify functional biomarkers in fNIRS signals of pain responses to noxious electrical stimuli, (2) the system is able to detect differences in pain responses across subjects and use this information to do a personalized analysis
and (3) we show that adopting a personalized approach to account for inter-individual variability significantly improves pain detection performance. The proof of principle approach used may provide a basis for deployment in patients with the ability to have real time differentiation of pain state.

\section{Data}

\subsection{Data acquisition}
\begin{figure}
	\centering
	\includegraphics[width=0.95\linewidth]{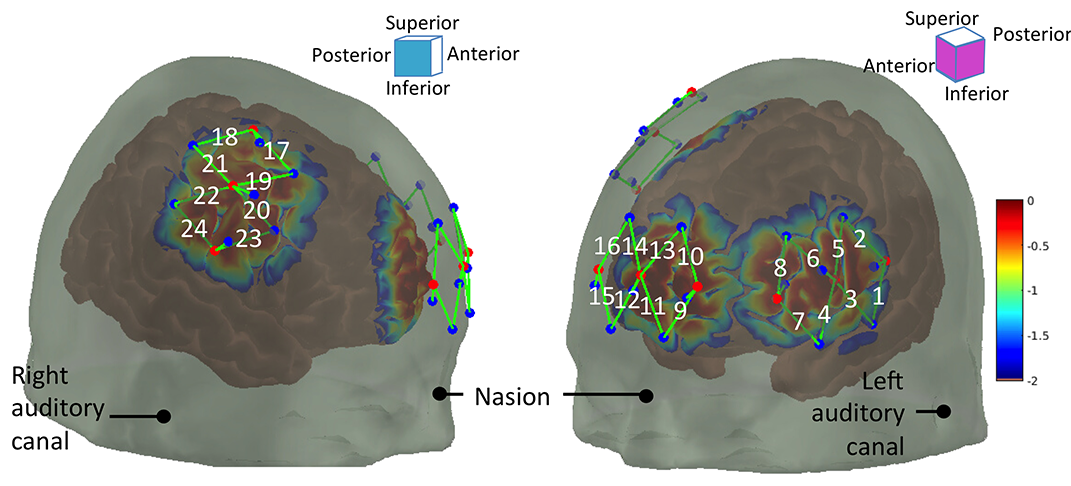}
	\caption{The arrangement of the fNIRS optodes and the corresponding sensitivity profile in mm$^{-1}$, showing the location of the detectors (blue dots), the sources (red dots) and the 24 channels (green lines).}
	\label{fig:mount}
\end{figure}

Twenty healthy subjects were recruited for this study (right handed, males, ages 19 to 38, mean age 28.7). Each subject gave written consent before the start of the scan. We used a multichannel continuous wave fNIRS system (CW7, TechEn, Massachusetts) operating at 690- and 830-nm wavelengths and sampling at 25 Hz with nine light emitters and twelve light detectors mounted on the subject’s head. These fNIRS optodes formed 24 30-mm channels and covered the anterior portion of the prefrontal cortex, the left dorsolateral prefrontal cortex and the right sensorimotor regions (see Fig.\ref{fig:mount}). An extra light detector was placed 8 mm from each emitter to create a short separation channel measuring the contribution of signals from extracerebral layers (e.g. skin, scalp and skull).

Pain was produced using an electrical stimulus.  Specifically, each subject’s level of subjective pain perception was pre-determined by applying 5 Hz electrical stimulations (Neurotron, Maryland) to his left thumb and asking him to report at a rating of 7 on a 0 to 10 scale, with the description of the sensation being “the subject should perceive pain but the pain should be just tolerable without breath holding or any retreat actions”, as well as at a rating of 3, corresponding to "the subject should be strongly aware of the stimulus but should not perceive any pain". The intensity of the current corresponding to the 3/10  and 7/10 sensation levels were used in the subsequent fNIRS sessions. 
Then, in the fNIRS sessions, we first recorded a pre-stimulation baseline scan, and then a stimulation scan in which we induced a randomized sequence of 6 painful electrical stimuli (rating: 7/10)  mixed with 6 innocuous electrical stimuli (nonpainful, rating: 3/10), all of which had a 5-second duration. The 12 stimuli were applied to the left thumb of a subject separated by 25 seconds of resting period. 

In total, we acquired 38 fNIRS sessions from 20 subjects: 7 subjects had one session, 8 subjects had two sessions, and 5 subjects had three sessions.
Next, we develop an automated binary classifier that takes fNIRS data windows as input, and learns to label the output as painful/not painful, based on whether the window arises from a painful electrical stimulus or a baseline segment.

\subsection{Data preprocessing}

The Matlab toolbox HOMER2 \cite{homer} was used to process the fNIRS data. As in \cite{Peng2018}, we first converted the recorded timecourses of the two wavelengths (690nm and 830nm) of light into optical density changes, and performed a 3rd order Butterworth low-pass filtering at a cutoff frequency of 0.5Hz. Based on the optical properties of both wavelengths, the optical density data were transformed into hemoglobin concentration changes (HbO and HbR) using the modified Beer-Lambert Law \cite{Delpy1988,Cope1988,Boas2004}. Following \cite{Peng2018}, the HbO and HbR timecourses of each channel were then processed under a general linear model (GLM), which included the timecourse of a short separation channel that showed the highest correlation with the current hemoglobin timecourse as a regressor to filter out the effect of physiological noises (such as heartbeat or respiration), as well as polynomial regressors up to the 3rd order to remove the trend in the data. The resulting HbO signal was used to extract machine learning features, as described in Sec.\ref{sec:features}.

\section{Methods}

\subsection{Feature extraction} \label{sec:features}

For each painful stimulus (rating: 7/10), we extracted windows of duration 20 seconds from the HbO signals for all 24 fNIRS channels, starting immediately after the onset of the stimulus. In addition to this, we randomly sampled "no pain" windows of same length from the pre-stimulation baseline scan. From these windows, we extracted (1) 10 b-spline coefficients and (2) 11 statistical features. 

\subsubsection{B-Spline coefficients} As in \cite{Pourshoghi2016}, we perform functional data analysis with cubic b-spline basis functions \cite{fdar} to extract $K$ basis coefficients from the 20 sec. windows to be used as features. Let's consider a window $\bm{w} = [w_1, ..., w_W]$ with $W$ datapoints. Our goal is to find the coefficients $\bm{c} = [c_1,...,c_K]$ such that 
\begin{equation}
s(t) \approx \sum_{k=1}^K c_k \phi_k (t) =  \bm{c}^\top \bm{\phi}
\end{equation}
where $\bm{\phi} = [\phi_1,...,\phi_K]$ is our cubic b-spline basis system. This is achieved by minimizing a penalized squared error cost function (PENSSE)
\begin{equation}
\texttt{PENSSE} = \underbrace{\sum_{i=1}^W [w_i - s(t_i)]^2}_{\texttt{SSE}} +  \lambda \underbrace{ \int \left [  \frac{d^2 s(t)}{dt^2} \right ]^2}_{\texttt{PEN}_2}
\end{equation}
where \texttt{SSE} is the sum of squared errors or residuals, and \texttt{PEN}$_2$ is the total curvature roughness penalty, which is introduced to provide smoothing. The weight of the roughness penalty relative to goodness of fit quantified in the \texttt{SSE} penalty is controlled by the smoothing parameter $\lambda$. As $\lambda$ increases, roughness becomes increasingly penalized. When $\lambda \rightarrow \infty$, we obtain an equivalent to a linear regression estimator. At the other extreme, when $\lambda \rightarrow 0$, the function $s(t)$ will overfit the data $\bm{w}$. To find the best value of $\lambda$, we can use the generalized cross-validation (GCV) measure \cite{Craven1978}
\begin{equation}
	\texttt{GCV}(\lambda) = \left (  \frac{W}{W-df(\lambda)} \right ) \left ( \frac{\texttt{SSE}}{W-df(\lambda)}  \right )
\end{equation}
where $df(\lambda)$ is the trace of the smoothing matrix $\bm{\mathrm{H}}(\lambda) = \bm{\phi} (\bm{\phi}^\top \bm{\phi} + \lambda \bm{\mathrm{R}} )^{-1} \bm{\phi}^\top$, and $\bm{\mathrm{R}} = \int \bm{\phi}(t) \bm{\phi}^\intercal (t) dt$ is the roughness penalty matrix \cite{fdar}.

B-spline functions $\phi_k (t) = B_k(t; m,\tau )$ are piecewise polynomials defined in terms of the order $m$ and the breakpoint or knot sequence $\tau_\ell$, $\ell =1,...,L-1$ where $L$ is the number of subintervals, and $k$ refers to the number of the largest knot at or to the immediate left of value $t$ \cite{fda}. Since the number of parameters required to define a spline function is the order plus the number of interior knots, we have that $K=m+L-1$.
In this work, we used cubic splines ($m=4$) and equally spaced knots, with $K=10$.

\subsubsection{Statistical features}
  In addition to the b-spline coefficients, we also calculated: (1) mean; (2) standard deviation; (3) maximum; (4) minimum,; (5) range; (6) the slope of the linear regression of $\bm{w}$ in its time series; (7) the location of maximum in the window $\bm{w}$, that is, ${\arg \max}_i \, (w_i) $; (8) the location of minimum in the window $\bm{w}$, that is, ${\arg \min}_i \, (w_i) $ ; (9) kurtosis; (10) skewness; and (11) area under the curve.

\subsection{Task assignment with spectral clustering} \label{sec:spectralclustering}

To assign each session $s$ to a task in our multi-task machine learning approach (Sec.\ref{sec:mtmkl}), we performed spectral clustering with the training features. As in \cite{dlmNIPS2017}, sessions within the same cluster were assigned to the same task.

For each session $s$ in our dataset, we build a descriptor vector $\bm{p}^{(s)} = [{p}^{(s)}_1,...,{p}^{(s)}_D]$ where $D$ is the dimensionality of our feature set, such that  $p_i^{(s)} = \frac{1}{N_{tr}^{(s)}} \sum_{j} x_i^{(s,j)}$, where $\bm{x}^{(s,j)} \in \mathcal{R}^D$ is $j$-th feature vector in the training set corresponding to session $s$, and $N_{tr}^{(s)}$ is the number of such feature vectors in that training set for session $s$.

\begin{figure}
	\centering
	\includegraphics[width=1\linewidth]{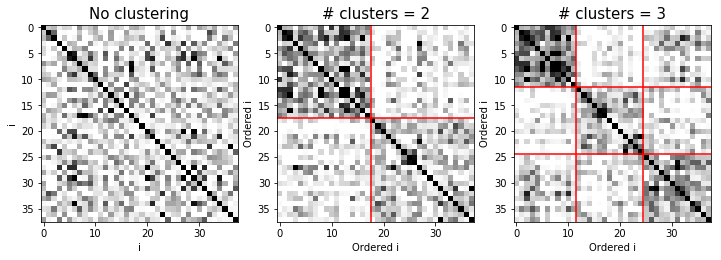}
	\caption{Task assignment using spectral clustering. The $i,j$ elements in the (clustered) similarity
matrices represent the distance from session $i$ to session $j$ in their average feature vector for label $y=+1$ (electrical pain 7/10). The darker the matrix element, the closer the corresponding subjects are.}
	\label{fig:clusters}
\end{figure}

We normalized these vectors to sum to 1 (to compensate for the scale differences), and used them to cluster sessions into different \textit{profiles}, each one corresponding to a task in our MT-ML algorithm. To this end, we used normalized spectral clustering~\cite{ShiMalik2000, Planck2006}. Namely, let $\bm{\hat{p}}^{(s)} $ be the normalized descriptor of session $s$. First, we construct a fully connected similarity graph and the corresponding weighted adjacency matrix $\bm{\mathrm{W}}$. We used the fully-connected graph, with edge weights $w_{ij}=K_{i,j} = K( \bm{\hat{p}}^{(i)} , \bm{\hat{p}}^{(j)} )$, and the radial basis function (RBF) kernel: $K(\bm{\hat{p}}^{(i)} , \bm{\hat{p}}^{(j)}) = \exp \left (-\gamma || \bm{\hat{p}}^{(i)} - \bm{\hat{p}}^{(j)} ||^2 \right )$ with $\gamma = 0.10$, as the similarity measure. 
Then, we build the degree matrix $\bm{\mathrm{D}}$ as the diagonal matrix with degrees $d_1,...,d_S$ on the diagonal, where $d_i$ is given by
$
d_i = \sum_{j=1}^{S} w_{ij}
$
where $S$ is the number of sessions in our dataset. Next, we compute the normalized graph Laplacian $\bm{\mathrm{L}} = \bm{\mathrm{I}} - \bm{\mathrm{D}}^{-1} \bm{\mathrm{W}}$ and calculate the first $T$ eigenvectors $\bm{u}_1,...,\bm{u}_T$ of $\bm{\mathrm{L}}$, where $T$ is the desired number of clusters. Let $\bm{\mathrm{U}} \in \mathbb{R}^{S\times T}$ be the matrix containing the vectors $\bm{u}_1,...,\bm{u}_T$ as columns. For $i=1,...,S$, let $\bm{y}_i \in \mathbb{R}^T$ be the vector corresponding to the $i$-th row of $\bm{\mathrm{U}}$. We cluster the points $(y_i)_{y=1,...,S}$ in $\mathbb{R}^T$ with the $k$-means algorithm into clusters $r_1,...,r_T$, where $T$ was determined by visual inspection of the grouped elements of $\bm{\mathrm{W}}$ after  clustering (see Fig.\ref{fig:clusters}).

\subsection{Multi-task multiple kernel machines} \label{sec:mtmkl}

In a single-task supervised learning scenario with training instances $\mathcal{D} = \{ (\bm{x}_i, y_i )\}_{i=1}^{N_{tr}}$, where $\bm{x}_i$ is a $D$-dimensional feature vector and $y_i \in \{ -1, 1 \}$ (no pain/pain), the goal is to learn a decision function $f$, such that $f(\bm{x_i}) \approx y_i$, that generalizes well in unseen data. 
For example, using the standard support vector machine (SVM) and the representer theorem, the decision function in the dual form becomes $f(\bm{x}) = \sum_{i=1}^{N_{tr}} \alpha_i k(\bm{x}_i,\bm{x}) + b$, where $N_{tr}$ is the training set size, $k: \mathbb{R}^D \times \mathbb{R}^D \rightarrow \mathbb{R} $ is the kernel function, and $\bm{\alpha}$ is the vector of Lagrange multipliers. 

Multiple kernel learning (MLT)  extends the traditional single kernel approach by allowing the introduction of a predefined set of kernels and  then letting the model learn an optimal linear combination of these kernels. By grouping the features extracted from each fNIRS channel $m$ according to channel, and assigning each of these feature groups a kernel, the model is able to automatically learn the importance of each of the channels to maximize the predictive accuracy \cite{Kandemir2014}. Briefly, each channel $m$
is represented by one kernel $k_m$, and then combined into a single kernel $k_\eta$ by the parametrization vector $\bm{\eta}$, such that
\begin{equation}
k_\eta (\bm{x}_i,\bm{x}_j; \bm{\eta}) = \sum_{m=1}^M \eta_m k_m (\bm{x}^{(m)}_i,\bm{x}^{(m)}_j)
\end{equation}
where $M=24$ is the number of fNIRS channels.

In MTL, we have a training set $\mathcal{D}^{(r)} = \{ (\bm{x}^{(r)}_i, y^{(r)}_i )\}_{i=1}^{N^{(r)}_{tr}}$ for each task $r$, and we are now interested in learning $T$ decision functions $f_r$, one for each task. The key idea behind MTL is that the models are learned simultaneously for all tasks. Information transfer between tasks is then achieved by imposing similarity between each $\bm{\eta}^{(r)}$ in the joint optimization problem. This way, the model is able to learn from all available data $\mathcal{D} = \cup_r \mathcal{D}^{(r)}$ while producing decision functions $f_r$ that are tailored to each task.

Multi-task multiple kernel learning (MT-MKL) \cite{Kandemir2014}, combines both concepts: (1) multiple kernel learning to account for different fNIRS channels, and (2) multi-task learning to personalize the models according to different clusters or tasks. Briefly, model parameters are learned by solving the following min-max optimization problem:

{ \footnotesize
\begin{equation} \label{eq:optimzation}
\underset{ \left \{ \bm{\eta}^{(r)} \in \mathcal{E} \right \}_{r=1}^{T}  }{\text{minimize}}
\underbrace{
\left \{ 
\underset{ \left \{ \bm{\alpha}^{(r)} \in \mathcal{A}^{(r)} \right \}_{r=1}^{T} }{\text{maximize}}  \Omega ( \{ \bm{\eta}^{(r)} \}_{r=1}^{T} )
+ \sum_{r=1}^T J^{(r)} (\bm{\alpha}^{(r)},\bm{\eta}^{(r)} )
\right \}
}_{\mathcal{O}_\eta}
\end{equation}
}
where $\Omega (\cdot )$ is the regularization term that imposes similarity between the kernels, $\mathcal{E} = \{ \bm{\eta} :\sum_{m=1}^{M} \eta_m = 1, \eta_m \geq 0 \; \forall m \}$ denotes the domain of the kernel combination parameters $\bm{\eta}^{(r)}$, $\mathcal{A}^{(r)}$ is the domain of the parameters of the kernel-based learner (the Lagrange multipliers) for task $r$, and $J^{(r)} (\cdot,\cdot)$ is the objective function of the kernel-based learner of task $r$.

In this work, as in \cite{Kandemir2014}, we consider two types of regularizers $\Omega (\cdot)$: (a) the $\ell_1$-norm regularizer
\begin{equation}
\Omega_1 ( \{ \bm{\eta}^{(r)} \}_{r=1}^T ) = -\nu \sum_{r=1}^T \sum_{s=1}^T  {\bm{\eta}^{(r)}}^\top \bm{\eta}^{(s)}
\end{equation}
and (b) the $\ell_2$-norm regularizer
\begin{equation}
\Omega_2 ( \{ \bm{\eta}^{(r)} \}_{r=1}^T ) = -\nu \sum_{r=1}^T \sum_{s=1}^T  || \bm{\eta}^{(r)} - \bm{\eta}^{(s)}   ||_2
\end{equation}
where the coefficient $\nu$ controls the influence of the regularizer on the cost function (Eq. \ref{eq:optimzation}). Hence, $\nu \rightarrow 0$ corresponds to treating the tasks as unrelated, whereas large values of $\nu$ enforces similar kernel combination parameters across the tasks $r$. 

\begin{algorithm}
\caption{Multitask Multiple Kernel Learning (MT-MKL)}\label{alg:mtmkl}
\begin{algorithmic}[1]
\State Initialize $\bm{\eta}^{(r)}$ as $(1/T, ..., 1/T)$, $\forall r$
\Repeat 
\State Calculate $\bm{K}^{(r)}_\eta = \{ k_\eta^{(r)}  (\bm{x}_i^{(r)}, \bm{x}_j^{(r)}; \bm{\eta}^{(r)}) \}^{N^{(r)}}_{i,j=1} $, $\forall r$
\State Solve a single-kernel machine using $\bm{K}^{(r)}_\eta$, $\forall r$
\State Update $\bm{\eta}^{(r)}$ in the opposite direction of $\partial \bm{\mathcal{O}}_n / \partial \bm{\eta}^{(r)}$, $\forall r$
\Until{convergence}
\end{algorithmic}
\end{algorithm}

The min-max optimization problem in Eq.\ref{eq:optimzation} can be solved using a two-step iterative  gradient descent algorihtm algorithm, summarized in Alg.\ref{alg:mtmkl} \cite{Kandemir2014}. As in \cite{Taylor2017}, we use a least-squares support vector machine (LSSVM) for each task-specific model. 

The gradient of the objective function $\mathcal{O}_n$ is

{ \small
\begin{equation}
\frac{\partial \mathcal{O}_n}{\partial \eta_m^{(r)}} = 
-2 \frac{\partial \bm{\Omega}(\bm{\eta}^{(r)})}{\partial \eta_m^{(r)}}
-\frac{1}{2} \sum_{i=1}^N \sum_{j=1}^N
\alpha_i^{(r)} \alpha_j^{(r)} y^{(r)} y^{(r)} k_m^{(r)} (\bm{x}_i^{(m)},\bm{x}_j^{(m)})
\end{equation}
}
where the gradients of the $\ell_1$-norm and  $\ell_2$-norm regularizers are
\begin{equation}
\frac{\partial \Omega_1(\bm{\eta}^{(r)})}{\partial \eta} = - \nu \sum_{s=1}^T \bm{\eta}^{(s)}
\end{equation}
\begin{equation}
\frac{\partial \Omega_2(\bm{\eta}^{(r)})}{\partial \eta} = - \nu \sum_{s=1}^T 2(\bm{\eta}^{(r)}- \bm{\eta}^{(s)})
\end{equation}

In this work, we consider two common types of kernels: the linear kernel $k(\bm{x}, \bm{x}') = \bm{x}^\top \bm{x}'$, and the radial basis function kernel $k(\bm{x}, \bm{x}') = \exp (- \gamma || \bm{x}-\bm{x}' ||^2)$.

\section{Results}

\begin{figure*}
	\centering
	\includegraphics[width=0.85\linewidth]{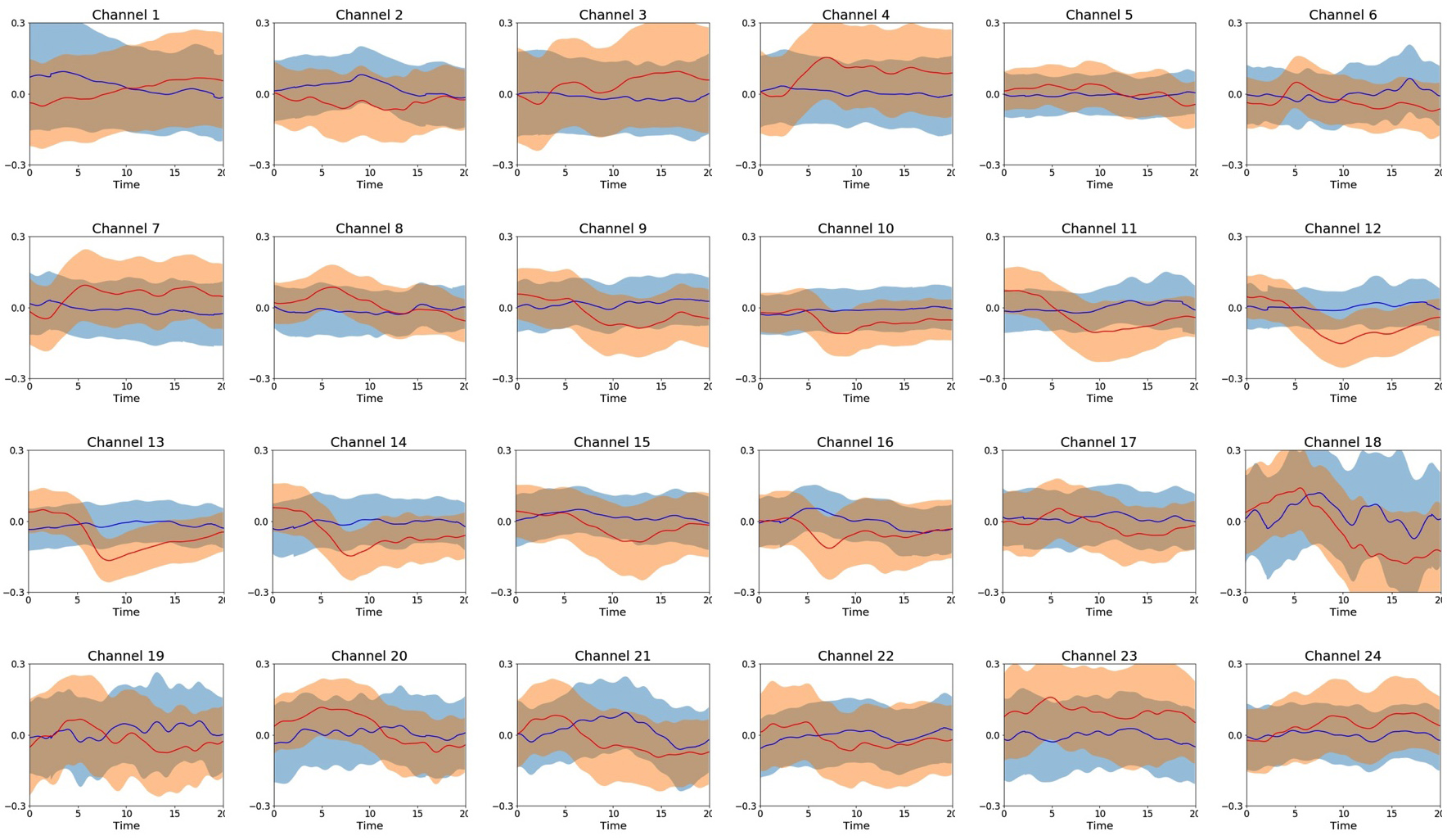} 
	\caption{fNIRS channels: comparison of the average HbO responses for 20 second windows randomly sampled from baseline (shown in blue) and extracted immediately after pain onset (shown in red). The thickness of the shadows represent half a standard deviation.}
	\label{fig:responses}
\end{figure*}

From our experimental recordings, we extracted windows of duration 20 seconds starting with the onset of the 7/10 electrical stimulus  as well as randomly sampled windows from the baseline recording . This choice of window size was done by visual inspection of the responses (see Fig.\ref{fig:responses}), and is in the same order of magnitude as those of other studies on pain detection from fNIRS signals \cite{Pourshoghi2016}. From these windows, we extracted 3 feature sets: (a) b-spline coefficients, (b) statistical features, and (c) b-spline coefficients and statistical features combined, that is, all features. 

First, we tested some standard single-task algorithms using 10-fold cross-validation on the balanced dataset. Data balancing was done by downsampling the over-represented class. The results are summarized in Table \ref{tab:stml}, and show a maximum accuracy of 72\% with the SVM\footnote{Please note that the SVMs in Table \ref{tab:stml} were trained with LIBSVM, whereas the $T=1$ SVMs in Table \ref{tab:mtmkl} were trained with our multikernel LSSVM algorithm.} with RBF kernel. 

Then, we investigated the effect of accounting for intersession variability via MTL with the MT-MKL algorithm described in Sec.\ref{sec:mtmkl}.
Following Sec.\ref{sec:spectralclustering}, for each session $s$ we calculated a descriptor vector $\bm{p}^{(s)}$ using the training data and one the three feature sets, and assigned each session to a cluster, for both $T=2$ and $T=3$, and also $T=1$ (no clustering, NC). Each cluster represented a task $r$ in our MT-MKL algorithm.
Then, we performed 10-fold cross-validation. Within the training set, we performed 5-fold cross-validation to choose the best combination of model hyperparameters.
Both $C$ and the regularization parameter $\nu$  were selected from the set $\{ 10^{-4},10^{-2},...,10^{+3} \}$. As before, when training the model, we balanced the training set by downsampling the over-represented class ($y=-1$). 

The results of the MT-MKL algorithm are shown in Table \ref{tab:mtmkl} and indicate that the b-spline coefficients tend to outperform both the statistical and combined features in classification performance. Furthermore, Table \ref{tab:mtmkl} shows a clear improvement in classification performance as we increase the number of clusters $T$. For all three feature sets, the best accuracy is achieved with $T=3$. The differences in accuracy with respect to the  $T=1$ and $T=2$ versions of the respective models are all statistically significant, with $p<0.05$ (2-tailed t-test). Among all model combinations, the best accuracy is obtained with $T=3$ and the RBF kernel with L1 regularization and the b-spline coefficients. 

Furthermore, we also examined the performance within clusters for the RBF L1 model with b-spline coefficients, for different values of $T$. 
The results are summarized in Table \ref{tab:insidelcusters} and indicate that our profiling approach finds clusters in which pain estimation performance is better, whereas other clusters seem to be more
challenging.

\begin{figure}
	\centering
	\includegraphics[width=0.8\linewidth]{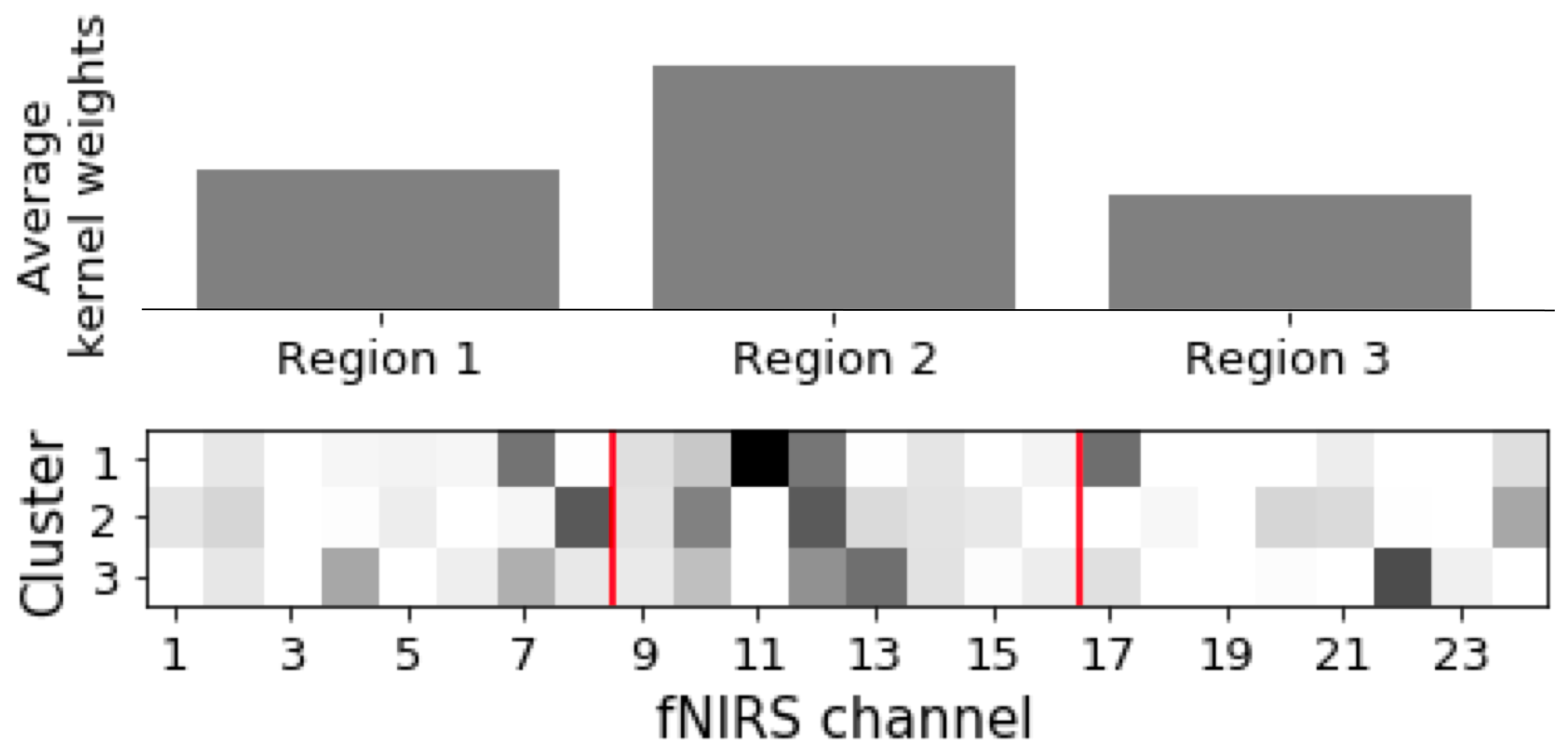}
	\caption{MT-MKL kernel weights ($\eta$). The bottom figure represents the kernel weights for each channel and cluster, i.e., task. The larger the  weights are, the darker the matrix elements are, representing increasing importance of that channel to the classifier for the corresponding binary classification task. The top figure shows the average kernel weights for each brain region, and indicates an increased importance of region 2 (channels 9-16), that is, the anterior prefrontal cortex (see Fig.\ref{fig:mount}). }
	\label{fig:kernelweights}
\end{figure}

Finally, we examined the kernel weights $\bm{\eta}$ produced by the $T=3$ RBF L1 model with b-spline coefficients. These are shown in Fig.\ref{fig:kernelweights} and indicate an increased importance of the anterior prefrontal cortex in classification performance. This result is consistent with a recent review which revealed a multidimensional role of the anterior prefrontal cortex in the process of pain perception \cite{Peng2017}.

\input{table_stml.tex}

\input{table_mtmkl.tex}

\input{table_insideclusters.tex}

\section{Conclusion}
In this study, we applied multi-task machine learning techniques to detect experimental electrical pain from fNIRS-measured brain signals in healthy male subjects. Using a MT-MKL approach to account for the inter-subject variability in pain response, we obtained an average detection accuracy of 80\% with the RBF kernel and B-spline coefficients. This study for the first time shows the potential of using fNIRS and machine learning techniques in providing an objective, automatic detection method of pain. Future work including more subjects (especially, with both genders) and other types of pain (e.g. spontaneous pain) may be needed to further validate this approach. 

\section*{Acknowledgements}
This work has been supported by the National Institute of Health (NIH)  National Institute of General Medical Sciences (NIGMS) grant (R01-GM122405) to DB. Daniel Lopez-Martinez is supported by MedImmune.


{\small
\bibliographystyle{IEEEtran}
\bibliography{Mendeley}
}

\end{document}

%% file: packages.tex
\usepackage[utf8]{inputenc} 
\usepackage[T1]{fontenc}    
\usepackage{hyperref}       
\usepackage{url}            
\usepackage{booktabs}       
\usepackage{amsfonts}       
\usepackage{nicefrac}       
\usepackage{microtype}      
\usepackage{graphics} 
\usepackage{epsfig} 
\usepackage{amsmath} 
\usepackage{amssymb}  
\usepackage{multirow}
\usepackage[draft]{todonotes} 
\usepackage{color}
\usepackage{bm}
\usepackage{verbatim}
\usepackage{algorithm}
\usepackage[noend]{algpseudocode}

\ifCLASSINFOpdf
\else
\fi

\makeatletter
\def\BState{\State\hskip-\ALG@thistlm}
\makeatother

%% file: table_stml.tex
\begin{table}[]
\centering
\caption{Classification performance of single-task models over 10-fold cross-validation.}
\label{tab:stml}
\begin{tabular}{lrrrr}
\hline
\textbf{Features \& model} & \multicolumn{1}{c}{\textbf{Accuracy}} & \multicolumn{1}{c}{\textbf{Sen.}} & \multicolumn{1}{c}{\textbf{Sp.}} & \multicolumn{1}{c}{\textbf{F$_1$}} \\ \hline
\textbf{Spline coefficients}   & \multicolumn{1}{l}{}     & \multicolumn{1}{l}{}     & \multicolumn{1}{l}{}  & \multicolumn{1}{l}{}                \\
Logistic regression (L1)       & 0.68      & 0.65         & 0.69              & 0.57              \\
Logistic regression (L2)       & 0.71            & 0.67                          & 0.73              & 0.60    \\
Linear Lasso            & 0.69              & 0.61              & 0.74        & 0.57      \\
SVM (linear kernel)           & 0.70     & 0.65        & 0.73     & 0.59          \\
SVM (rbf kernel)         & 0.72        & 0.81        & 0.67         & 0.65        \\
\textbf{Statistics}     & \multicolumn{1}{l}{}    & \multicolumn{1}{l}{}    & \multicolumn{1}{l}{}     & \multicolumn{1}{l}{}                \\
Logistic regression (L1)           & 0.63        & 0.64      & 0.62                  & 0.54         \\
Logistic regression (L2)           & 0.63      & 0.64         & 0.63          & 0.53        \\
Linear Lasso               & 0.65         & 0.64            & 0.66       & 0.54                   \\
SVM (linear kernel)       & 0.64       & 0.64     & 0.64      & 0.53             \\
SVM (rbf kernel)           & 0.65         & 0.76     & 0.60             & 0.58            \\
\textbf{All features}    & \multicolumn{1}{l}{}    & \multicolumn{1}{l}{}    & \multicolumn{1}{l}{}   & \multicolumn{1}{l}{}       \\
Logistic regression            & 0.70          & 0.67                    & 0.71             & 0.59         \\
Logistic regression (L2)        & 0.71               & 0.72             & 0.70                   & 0.6                \\
Linear Lasso           & 0.71         & 0.70             & 0.71                   & 0.61              \\
SVM (linear kernel)         & 0.71       & 0.72                   & 0.70        & 0.62          \\
SVM (rbf kernel)            & 0.71           & 0.84         & 0.65         & 0.66      
\end{tabular}
\end{table}

%% file: table_mtmkl.tex
\begin{table*}[]
\centering
\caption{Average accuracy (ACC), sensitivity (Sen.), specificity (Sp.) and F$_1$ score of the MT-MKL models over 10-fold cross-validation. The highest accuracy obtained with each feature set is highlighted in bold.}
\label{tab:mtmkl}
\small
\begin{tabular}{llllllllllllllllll}
\hline
\multicolumn{3}{l}{\textbf{MT-MKL model}}  &  & \multicolumn{4}{l}{\textbf{B-spline coefficients}} &  & \multicolumn{4}{l}{\textbf{Statistical features}} &  & \multicolumn{4}{l}{\textbf{All features}} \\ \cline{1-3} \cline{5-8} \cline{10-13} \cline{15-18} 
T                  & Kernel & Reg &  & ACC       & Sen.      & Sp.        & F1      &  & ACC      & Sen.      & Sp.        & F1      &  & ACC     & Sen.     & Sp.        & F1  \\ \hline
\multirow{4}{*}{1} & Linear & L1  &  & 0.68      & 0.74      & 0.66       & 0.60    &  & 0.65     & 0.68      & 0.63       & 0.56    &  & 0.72    & 0.77     & 0.71       & 0.64   \\
                   & Linear & L2  &  & 0.70      & 0.71      & 0.70       & 0.61    &  & 0.66     & 0.69      & 0.64       & 0.56    &  & 0.72    & 0.75     & 0.70       & 0.63   \\
                   & RBF    & L1  &  & 0.74      & 0.75      & 0.74       & 0.65    &  & 0.70     & 0.77      & 0.67       & 0.62    &  & 0.74    & 0.76     & 0.74       & 0.66   \\
                   & RBF    & L2  &  & 0.73      & 0.79      & 0.71       & 0.66    &  & 0.69     & 0.72      & 0.68       & 0.61    &  & 0.74    & 0.76     & 0.74       & 0.65   \\ \hline
\multirow{4}{*}{2} & Linear & L1  &  & 0.74      & 0.77      & 0.72       & 0.66    &  & 0.65     & 0.70      & 0.64       & 0.56    &  & 0.71    & 0.71     & 0.71       & 0.62   \\
                   & Linear & L2  &  & 0.74      & 0.75      & 0.73       & 0.65    &  & 0.64     & 0.65      & 0.63       & 0.54    &  & 0.71    & 0.73     & 0.70       & 0.62   \\
                   & RBF    & L1  &  & 0.77      & 0.79      & 0.75       & 0.69    &  & 0.70     & 0.76      & 0.67       & 0.62    &  & 0.72    & 0.79     & 0.69       & 0.65   \\
                   & RBF    & L2  &  & 0.76      & 0.76      & 0.75       & 0.68    &  & 0.70     & 0.77      & 0.67       & 0.63    &  & 0.73    & 0.73     & 0.73       & 0.64   \\ \hline
\multirow{4}{*}{3} & Linear & L1  &  & 0.75      & 0.76      & 0.74       & 0.66    &  & 0.68     & 0.70      & 0.67       & 0.59    &  & 0.75    & 0.75     & 0.75       & 0.66   \\
                   & Linear & L2  &  & 0.75      & 0.79      & 0.74       & 0.68    &  & 0.69     & 0.67      & 0.70       & 0.59    &  & 0.76    & 0.76     & 0.75       & 0.67   \\
                   & RBF    & L1  &  & \textbf{0.80}      & 0.83      & 0.78       & 0.73    &  & 0.72     & 0.78      & 0.68       & 0.65    &  & 0.76    & 0.81     & 0.74       & 0.69   \\
                   & RBF    & L2  &  & 0.79      & 0.85      & 0.76       & 0.72    &  & \textbf{0.73}     & 0.75      & 0.72       & 0.64    &  & \textbf{0.77}    & 0.81     & 0.75      & 0.70   \\ \hline
\end{tabular}
\end{table*}

%% file: table_insideclusters.tex
\begin{table}[]
\centering
\caption{Average performance of the MT-MKL RBF L1 model over 10-fold cross-validation, for different values of $T$, corresponding to the clusters shown in Figure \ref{fig:clusters}}
\label{tab:insidelcusters}
\begin{tabular}{|c|c|c|c|c|c|}
\hline
\textbf{\# clusters} & \textbf{Cluster} & \textbf{Acc.} & \textbf{Sen.} & \textbf{Sp.} & \textbf{F1} \\ \hline\hline
NC (T=1)                  & 1                & 0.74                 & 0.75                    & 0.74                    & 0.65                 \\ \hline \hline
\multirow{2}{*}{T=2} & 1                & 0.81                 & 0.87                    & 0.77                    & 0.74                 \\ \cline{2-6} 
                     & 2                & 0.67                 & 0.77                    & 0.61                    & 0.60                 \\ \hline \hline
\multirow{3}{*}{T=3} & 1                & 0.87                 & 0.94                    & 0.83                    & 0.82                 \\ \cline{2-6} 
                     & 2                & 0.73                 & 0.82                    & 0.69                    & 0.66
                 \\ \cline{2-6} 
                     & 3                & 0.80                 & 0.88                    & 0.77                    & 0.74                 \\ \hline
\end{tabular}
\end{table}

%% file: root.bbl
\begin{thebibliography}{10}
\providecommand{\url}[1]{#1}
\csname url@samestyle\endcsname
\providecommand{\newblock}{\relax}
\providecommand{\bibinfo}[2]{#2}
\providecommand{\BIBentrySTDinterwordspacing}{\spaceskip=0pt\relax}
\providecommand{\BIBentryALTinterwordstretchfactor}{4}
\providecommand{\BIBentryALTinterwordspacing}{\spaceskip=\fontdimen2\font plus
\BIBentryALTinterwordstretchfactor\fontdimen3\font minus
  \fontdimen4\font\relax}
\providecommand{\BIBforeignlanguage}[2]{{%
\expandafter\ifx\csname l@#1\endcsname\relax
\typeout{** WARNING: IEEEtran.bst: No hyphenation pattern has been}%
\typeout{** loaded for the language `#1'. Using the pattern for}%
\typeout{** the default language instead.}%
\else
\language=\csname l@#1\endcsname
\fi
#2}}
\providecommand{\BIBdecl}{\relax}
\BIBdecl

\bibitem{DeCWilliams2016}
A.~C. d.~C. Williams and K.~D. Craig, ``{Updating the definition of pain},''
  \emph{PAIN}, vol. 157, no.~11, pp. 2420--2423, 11 2016.

\bibitem{Younger2009}
J.~Younger, R.~McCue, and S.~Mackey, ``{Pain outcomes: A brief review of
  instruments and techniques},'' \emph{Current Pain and Headache Reports},
  vol.~13, no.~1, pp. 39--43, 2009.

\bibitem{dlm_MTL_2017}
D.~Lopez-Martinez and R.~Picard, ``{Multi-task Neural Networks for Personalized
  Pain Recognition from Physiological Signals},'' in \emph{Seventh
  International Conference on Affective Computing and Intelligent Interaction
  Workshops and Demos (ACIIW)}, San Antonio, TX, 2017.

\bibitem{dlmNIPS2017}
D.~Lopez-Martinez, O.~Rudovic, and R.~Picard, ``{Physiological and Behavioral
  Profiling for Nociceptive Pain Estimation Using Personalized Multitask
  Learning.}'' in \emph{Neural Information Processing Systems (NIPS) Workshop
  on Machine Learning for Health}, Long Beach, USA, 2017.

\bibitem{LopezMartinez2017c}
\BIBentryALTinterwordspacing
------, ``{Personalized Automatic Estimation of Self-Reported Pain Intensity
  from Facial Expressions},'' in \emph{2017 IEEE Conference on Computer Vision
  and Pattern Recognition Workshops (CVPRW)}.\hskip 1em plus 0.5em minus
  0.4em\relax Hawaii, USA: IEEE, 7 2017, pp. 2318--2327.
\BIBentrySTDinterwordspacing

\bibitem{Aasted2016}
C.~M. Aasted, M.~A. Yucel, S.~C. Steele, K.~Peng, D.~A. Boas, L.~Becerra, and
  D.~Borsook, ``{Frontal lobe hemodynamic responses to painful stimulation: A
  potential brain marker of nociception},'' \emph{PLoS ONE}, vol.~11, no.~11,
  pp. 1--12, 2016.

\bibitem{Gelinas2010}
\BIBentryALTinterwordspacing
C.~G{\'{e}}linas, M.~Choini{\`{e}}re, M.~Ranger, A.~Denault, A.~Deschamps, and
  C.~Johnston, ``{Toward a new approach for the detection of pain in adult
  patients undergoing cardiac surgery: Near-infrared spectroscopy—A pilot
  study},'' \emph{Heart {\&} Lung: The Journal of Acute and Critical Care},
  vol.~39, no.~6, pp. 485--493, 2010.
\BIBentrySTDinterwordspacing

\bibitem{Yucel2015}
\BIBentryALTinterwordspacing
M.~A. Y{\"{u}}cel, C.~M. Aasted, M.~P. Petkov, D.~Borsook, D.~A. Boas, and
  L.~Becerra, ``{Specificity of hemodynamic brain responses to painful stimuli:
  a functional near-infrared spectroscopy study.}'' \emph{Scientific reports},
  vol.~5, p. 9469, 2015.
\BIBentrySTDinterwordspacing

\bibitem{Obrig2014}
\BIBentryALTinterwordspacing
H.~Obrig, ``{NIRS in clinical neurology — a ‘promising’ tool?}''
  \emph{NeuroImage}, vol.~85, pp. 535--546, 1 2014.
\BIBentrySTDinterwordspacing

\bibitem{Pourshoghi2016}
\BIBentryALTinterwordspacing
A.~Pourshoghi, I.~Zakeri, and K.~Pourrezaei, ``{Application of functional data
  analysis in classification and clustering of functional near-infrared
  spectroscopy signal in response to noxious stimuli},'' \emph{Journal of
  Biomedical Optics}, vol.~21, no.~10, 2016.
\BIBentrySTDinterwordspacing

\bibitem{Dube2009}
A.-A. Dub{\'{e}}, M.~Duquette, M.~Roy, F.~Lepore, G.~Duncan, and P.~Rainville,
  ``{Brain activity associated with the electrodermal reactivity to acute heat
  pain},'' \emph{NeuroImage}, vol.~45, no.~1, pp. 169--180, 3 2009.

\bibitem{Caruana1997}
R.~Caruana, ``{Multitask Learning},'' \emph{Machine Learning}, vol.~28, no.~1,
  pp. 41--75, 1997.

\bibitem{Planck2006}
U.~von Luxburg, ``{A tutorial on spectral clustering},'' \emph{Statistics and
  Computing}, vol.~17, no.~4, pp. 395--416, 12 2007.

\bibitem{Kandemir2014}
\BIBentryALTinterwordspacing
M.~Kandemir, A.~Vetek, M.~G{\"{o}}nen, A.~Klami, and S.~Kaski, ``{Multi-task
  and multi-view learning of user state},'' \emph{Neurocomputing}, vol. 139,
  pp. 97--106, 9 2014.
\BIBentrySTDinterwordspacing

\bibitem{homer}
\BIBentryALTinterwordspacing
T.~J. Huppert, S.~G. Diamond, M.~A. Franceschini, and D.~A. Boas, ``{HomER: a
  review of time-series analysis methods for near-infrared spectroscopy of the
  brain.}'' \emph{Applied optics}, vol.~48, no.~10, pp. 280--98, 4 2009.
\BIBentrySTDinterwordspacing

\bibitem{Peng2018}
K.~Peng, M.~A. Y{\"{u}}cel, C.~M. Aasted, S.~C. Steele, D.~A. Boas, D.~Borsook,
  and L.~Becerra, ``{Using prerecorded hemodynamic response functions in
  detecting prefrontal pain response: a functional near-infrared spectroscopy
  study.}'' \emph{Neurophotonics}, vol.~5, no.~1, 1 2018.

\bibitem{Delpy1988}
\BIBentryALTinterwordspacing
D.~T. Delpy, M.~Cope, P.~V.~D. Zee, S.~Arridge, S.~Wray, and J.~Wyatt,
  ``{Estimation of optical pathlength through tissue from direct time of flight
  measurement},'' \emph{Physics in Medicine and Biology}, vol.~33, no.~12, pp.
  1433--1442, 12 1988.
\BIBentrySTDinterwordspacing

\bibitem{Cope1988}
\BIBentryALTinterwordspacing
M.~Cope and D.~T. Delpy, ``{System for long-term measurement of cerebral blood
  and tissue oxygenation on newborn infants by near infra-red
  transillumination.}'' \emph{Medical {\&} biological engineering {\&}
  computing}, vol.~26, no.~3, pp. 289--94, 5 1988.
\BIBentrySTDinterwordspacing

\bibitem{Boas2004}
D.~A. Boas, A.~M. Dale, and M.~A. Franceschini, ``{Diffuse optical imaging of
  brain activation: Approaches to optimizing image sensitivity, resolution, and
  accuracy},'' \emph{NeuroImage}, vol.~23, no. SUPPL. 1, 2004.

\bibitem{fdar}
\BIBentryALTinterwordspacing
J.~Ramsay, G.~Hooker, and S.~Graves, \emph{{Functional Data Analysis with R and
  MATLAB}}.\hskip 1em plus 0.5em minus 0.4em\relax New York, NY: Springer New
  York, 2009.
\BIBentrySTDinterwordspacing

\bibitem{Craven1978}
\BIBentryALTinterwordspacing
P.~Craven and G.~Wahba, ``{Smoothing noisy data with spline functions},''
  \emph{Numerische Mathematik}, vol.~31, no.~4, pp. 377--403, 12 1978.
\BIBentrySTDinterwordspacing

\bibitem{fda}
\BIBentryALTinterwordspacing
J.~Ramsay and B.~W. Silverman, \emph{{Functional Data Analysis}}, ser. Springer
  Series in Statistics.\hskip 1em plus 0.5em minus 0.4em\relax Springer-Verlag,
  2005.
\BIBentrySTDinterwordspacing

\bibitem{ShiMalik2000}
\BIBentryALTinterwordspacing
{Jianbo Shi} and J.~Malik, ``{Normalized cuts and image segmentation},''
  \emph{IEEE Transactions on Pattern Analysis and Machine Intelligence},
  vol.~22, no.~8, pp. 888--905, 2000.
\BIBentrySTDinterwordspacing

\bibitem{Taylor2017}
S.~A. Taylor, N.~Jaques, E.~Nosakhare, A.~Sano, and R.~Picard, ``{Personalized
  Multitask Learning for Predicting Tomorrow's Mood, Stress, and Health},''
  \emph{IEEE Transactions on Affective Computing}, 2017.

\bibitem{Peng2017}
\BIBentryALTinterwordspacing
K.~Peng, S.~C. Steele, L.~Becerra, and D.~Borsook, ``{Brodmann area 10:
  Collating, integrating and high level processing of nociception and pain},''
  \emph{Progress in Neurobiology}, 2017.
\BIBentrySTDinterwordspacing

\end{thebibliography}
